\documentclass[10pt,twocolumn,letterpaper]{article}

\usepackage{iccv}
\usepackage{times}
\usepackage{epsfig}
\usepackage{graphicx}
\usepackage{amsmath}
\usepackage{amssymb}
\usepackage{tabularray}
\usepackage{booktabs}
\usepackage{tabularx}
\usepackage{float}
\usepackage{siunitx}
\usepackage[table]{xcolor}
\usepackage{hhline}
\usepackage[margin=2cm]{geometry}
\usepackage{adjustbox}
\usepackage{multirow}
\usepackage{array}
\usepackage{boldline} 
\usepackage{makecell}
\usepackage{xcolor}
\usepackage{pifont}
\newcommand{\xmark}{\ding{55}}

\usepackage[pagebackref=true,breaklinks=true,letterpaper=true,colorlinks,bookmarks=false]{hyperref}

\iccvfinalcopy % *** Uncomment this line for the final submission

\def\iccvPaperID{****} % *** Enter the ICCV Paper ID here

\begin{document}

%%%%%%%%% TITLE
\title{Self-supervised Semantic Segmentation: Consistency over Transformation}

% \author{Sanaz Karimijafarbigloo$^{\;1}$
% \quad
% Reza Azad$^{\;2}$
% \quad
% Amirhossein Kazerouni$^{\;3}$
% \quad
% Yury Velichko$^{\;4}$ \\
% \quad
% Ulas Bagci$^{\;4}$
% \quad
% Dorit Merhof$^{\;1}$
% \\
% ${^1}$University of Regensburg, Germany
% \quad
% ${^2}$RWTH Aachen University, Germany
% \quad 
% \\
% ${^3}$Iran University of Science and Technology, Iran
% \quad 
% ${^4}$Northwestern University, USA
% \\
% {\tt\small \{sanaz.karimijafarbigloo, dorit.merhof\}@ur.de, \{rezazad68, amirhossein477\}@gmail.com} \\
% {\tt\small  \{ulas.bagci, y-velichko\}@northwestern.edu}
% }

\author{Sanaz Karimijafarbigloo$^{\;1}$
\quad
Reza Azad$^{\;2}$
\quad
Amirhossein Kazerouni$^{\;3}$
\quad
Yury Velichko$^{\;4}$ \\
\quad
Ulas Bagci$^{\;4}$
\quad
Dorit Merhof$^{\;1,5}$
\\
${^1\;}$Faculty of Informatics and Data Science, University of Regensburg, Germany
\\
${^2\;}$Faculty of Electrical Engineering and Information Technology, RWTH Aachen University, Germany
\\
${^3\;}$School of Electrical Engineering, Iran University of Science and Technology, Iran
\\
${^4\;}$Department of Radiology, Northwestern University, Chicago, USA
\\
${^5\;}$ Fraunhofer Institute for Digital Medicine MEVIS, Bremen, Germany
\\
{\tt\small \{sanaz.karimijafarbigloo, dorit.merhof\}@ur.de, \{rezazad68, amirhossein477\}@gmail.com} \\
{\tt\small  \{ulas.bagci, y-velichko\}@northwestern.edu}
}

% \and

\maketitle
\ificcvfinal\thispagestyle{empty}\fi

%%%%%%%%% ABSTRACT
\begin{abstract}
Accurate medical image segmentation is of utmost importance for enabling automated clinical decision procedures. However, prevailing supervised deep learning approaches for medical image segmentation encounter significant challenges due to their heavy dependence on extensive labeled training data. To tackle this issue, we propose a novel self-supervised algorithm, \textbf{S$^3$-Net}, which integrates a robust framework based on the proposed Inception Large Kernel Attention (I-LKA) modules. This architectural enhancement makes it possible to comprehensively capture contextual information while preserving local intricacies, thereby enabling precise semantic segmentation. Furthermore, considering that lesions in medical images often exhibit deformations, we leverage deformable convolution as an integral component to effectively capture and delineate lesion deformations for superior object boundary definition. Additionally, our self-supervised strategy emphasizes the acquisition of invariance to affine transformations, which is commonly encountered in medical scenarios. This emphasis on robustness with respect to geometric distortions significantly enhances the model's ability to accurately model and handle such distortions. To enforce spatial consistency and promote the grouping of spatially connected image pixels with similar feature representations, we introduce a spatial consistency loss term. This aids the network in effectively capturing the relationships among neighboring pixels and enhancing the overall segmentation quality. The S$^3$-Net approach iteratively learns pixel-level feature representations for image content clustering in an end-to-end manner. Our experimental results on skin lesion and lung organ segmentation tasks show the superior performance of our method compared to the SOTA approaches. \href{https://github.com/mindflow-institue/SSCT}{\textcolor{magenta}{Github}}.

\end{abstract}

\section{Introduction}

Over the past decade, deep learning approaches have achieved significant success, which can largely be attributed to the progress made in supervised learning research. 
However, the efficacy of these methods is highly dependent on the availability of a large amount of annotated training data. In situations where annotated data is limited or resource-intensive to obtain, these approaches may prove inefficient. One domain where this scarcity is evident in medical image analysis. Given the large size of medical images and the importance of precise labeling, which requires experts, the process of providing a wide range of manually annotated data is time-consuming, labor-intensive, and expensive \cite{azad2022medical,jose2023end,molaei2023implicit}. Moreover, the process of manual segmentation and labeling is prone to human error. To mitigate the labor-intensive nature of annotation, several strategies have been proposed in the literature.  One such strategy is transfer learning, which serves as a benchmark approach. Transfer learning facilitates the process of representational learning by fine-tuning the pre-trained network for the new task at hand. While knowledge transfer provides a promising starting point for the optimization algorithm, the scarcity of annotated data on the downstream task limits the network's convergence and its ability to learn task-specific features, resulting in less stable models. Moreover, in complex tasks such as segmentation, this approach proves to be inefficient due to the predefined model architecture \cite{araujo2022automatic, alhares2023amtldc,mahbod2020transfer}. Unsupervised methods offer an alternative solution by reformulating the problem based on learning features directly from the data itself \cite{ gao2022large,melas2022deep, hamilton2022unsupervised,liu2023memory,karimijafarbigloo2023self,ahn2020unsupervised}. However, the reliability of these approaches is not always guaranteed as no label or metric is available to validate their effectiveness.

A semi-supervised algorithm, which is a machine learning methodology that combines labeled and unlabeled data to construct predictive models, is also another approach to tackle the problem of data scarcity. The labeled data provides explicit supervision, guiding the learning process, while the unlabeled data contributes additional information for capturing underlying patterns and data structure \cite{luo2021semi,luo2022semi,you2023rethinking}. Nevertheless, the effectiveness of semi-supervised approaches can be compromised when the labeled data fails to adequately represent the entire distribution. Furthermore, although semi-supervised learning reduces the need for extensive manual labeling, it still necessitates a small set of labeled data. The process of annotation, even on a smaller scale, can be time-consuming, expensive, and dependent on domain expertise. The high cost associated with labeling data may hinder the scalability and practicality of semi-supervised methods. Additionally, labeling bias is another limitation of this approach.

In contrast to the previously mentioned strategies that rely on modeling the data distribution, the self-supervised technique has attracted recent interest and uses a different perspective by introducing a set of matching tasks \cite{caron2020unsupervised,chen2020simple,doersch2015unsupervised,kalapos2022self}. SSL has gained significant acceptance as a viable technique for learning medical image representations without specialized annotations. This approach allows for learning semantic features by generating supervisory signals from a large set of unlabeled data, effectively eliminating the need for human annotation \cite{chen2019self}. SSL consists of two main tasks: the pretext task and the downstream task. In the pretext task, where the SSL takes place, a model is trained in a supervised manner using unlabeled data. Labels are generated from the data to guide the model to learn semantic segmentation from the data. Subsequently, the learned representations obtained from the pretext task are transferred to the downstream task as initial weights. This transfer of weights allows the model to fine-tune and successfully achieve its intended goal.

Contrastive learning (CL), which has been extensively studied by various researchers \cite{zbontar2021barlow,haghighi2022dira,fischer2023self}, is a successful variant of SSL that has the ability to achieve the performance of SOTA algorithms even with a small number of annotated data. The CL methods aim to increase the similarity between representations of differently augmented input samples (referred to as positive pairs) while ensuring that representations of distinct samples (referred to as negatives) are dissimilar. The resulting neural network parameters are well-suited for initializing downstream tasks, where the learned representations from the pretext task are fine-tuned to adapt to the specific downstream task. This approach has been extended to handle dense pixel-wise image data, facilitating semantic segmentation even with limited available data \cite{liu2022improving,he2022intra,chaitanya2023local}.  

Despite the promising results achieved by CL, we argue that some aspects are relatively unexplored in the existing literature. Addressing these gaps has the potential to improve the current SOTA methods in medical imaging.
The first limitation arises when the unlabeled dataset used for training self-supervised contrastive learning contains biases or imbalances. In such scenarios, the learned representations may inherit these biases, as the learned representations are derived only from the intrinsic structure and patterns in the unlabeled data, resulting in biased predictions or limited generalization capabilities. 
Second, accurate semantic segmentation requires a model that can effectively capture long-range dependencies and maintain local consistency within images. Third, in the medical domain, it is important to consider that lesions often exhibit deformations in their shapes. Therefore, a learning algorithm employed for medical image analysis should possess the capability to capture and understand such deformations. Moreover, the algorithm should be invariant to common transformations encountered in medical images, such as shear, scale, and rotation. This ensures that the algorithm can effectively handle variations and changes in the appearance of lesions.

\begin{figure*}[t]
	\centering
	\includegraphics[width=\textwidth]{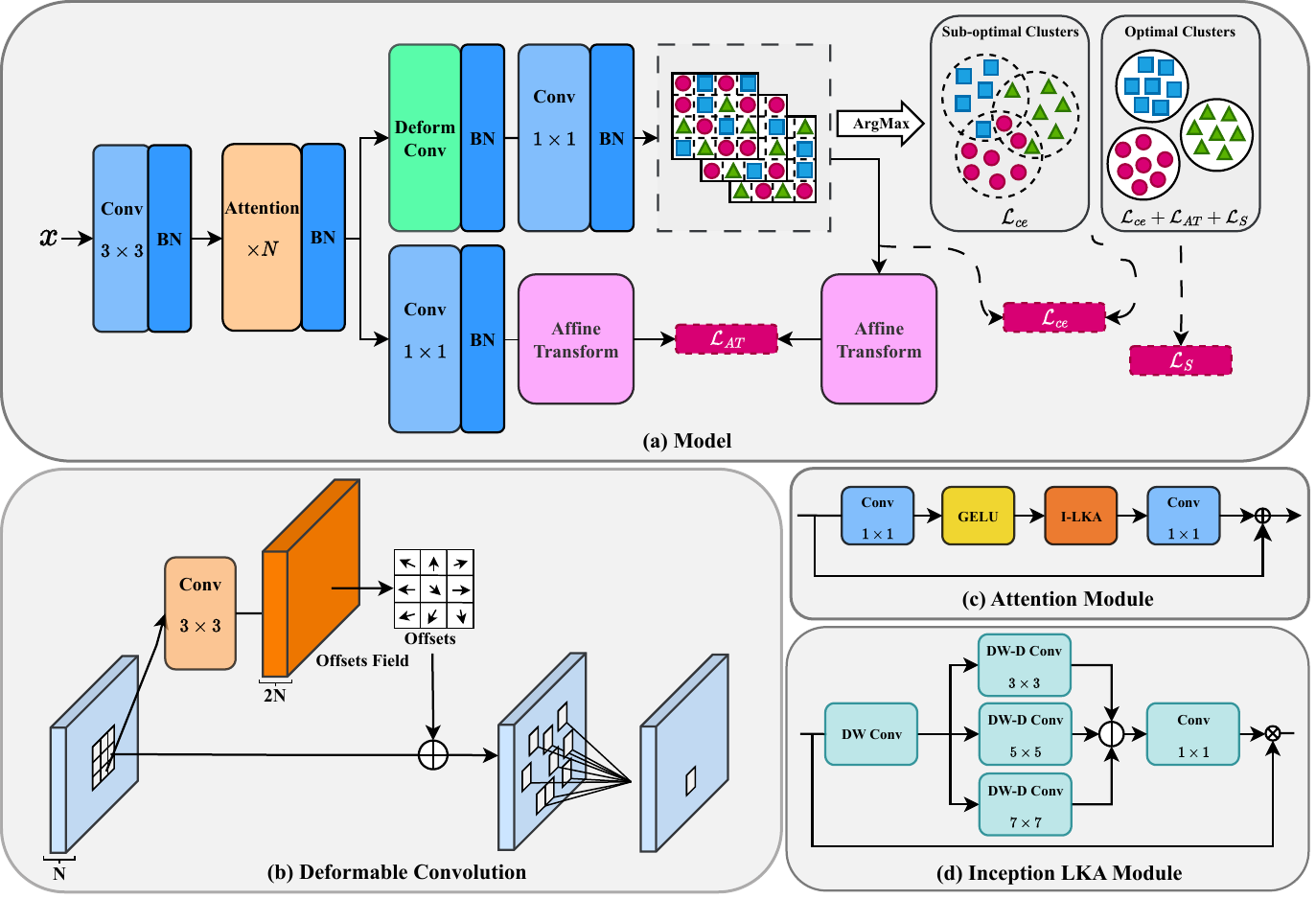}
	\caption{A general overview of the S$^3$-Net framework. The I-LKA module captures the intricate relationships between local and global features. Subsequently, it leverages a self-supervised learning strategy to enforce feature consistency throughout the network's predictions.}
    
	\label{fig:Model}
\end{figure*}

To address the encountered challenges outlined above, first, \ding{182} we propose I-LKA modules, which serve as a fundamental building block in our network design. These modules are specifically designed to capture contextual information comprehensively while preserving local descriptions. By striking a balance between these two aspects, our architecture facilitates precise semantic segmentation by effectively leveraging both global and local information.
Recognizing the prevalence of deformations in medical image lesions, \ding{183} we incorporate deformable convolution as a crucial component in our approach. This enables our model to effectively capture and delineate deformations, leading to improved boundary definition for the identified objects.
In order to make our model more robust to geometric transformations commonly encountered in medical scenarios, \ding{184} we integrate a self-supervised algorithm based on contrastive learning. By emphasizing the acquisition of invariance to affine transformations, our approach enhances the model's capacity to handle such transformations effectively. This allows the model to better generalize and adapt to different spatial configurations and orientations, thus improving overall performance in medical image segmentation tasks.
To ensure spatial consistency and promote the grouping of spatially connected pixels with similar features, \ding{185} we model a spatial consistency loss term based on edge information. This loss term facilitates the learning process by encouraging the network to capture the relationships among neighboring pixels.
Finally, \ding{186} our proposed method (\autoref{fig:Model}) effectively tackles dataset bias by performing the prediction process based on a single image only. This approach helps to mitigate the potential bias that may arise from imbalanced or skewed datasets.
\section{Related Works}
SSL has shown significant benefits for vision tasks by allowing the extraction of semantic and effective representations from unlabeled data in an unsupervised manner. This approach is rooted in the idea that significant performance improvements can be achieved by enhancing representation learning. A specific variant of SSL, known as contrastive learning, has gained substantial attention in recent years. Contrastive learning approaches strive to acquire resilient representations by optimizing similarity constraints, enabling the discrimination between similar (positive) and dissimilar (negative) pairs within a given dataset. In this direction, Chaitanya et al. \cite{chaitanya2020contrastive} introduced a contrastive learning framework specifically designed for segmenting volumetric medical images in a semi-supervised scenario. Their approach involved leveraging contrasting strategies that take advantage of the structural similarity present in volumetric medical images. Additionally, they incorporated a local variant of the contrastive loss to learn distinctive representations of local regions that are useful for per-pixel segmentation. You et al. \cite {you2022simcvd} introduced SimCVD, a contrastive distillation framework that improves voxel-wise representation learning. This contrastive training strategy involves using two views of an input volume and predicting their signed distance maps, using only two independent dropout masks. Additionally, they performed structural distillation by distilling pair-wise similarities. In another work, Moriya et al. \cite{moriya2018unsupervised} suggested the utilization of k-means clustering for grouping pixels with similar characteristics in micro-CT images, facilitating the extraction of significant feature representations. Nonetheless, self-supervised clustering methods encounter constraints such as the need for manual cluster size selection and difficulties posed by complex regions characterized by fuzzy boundaries, diverse shapes, artifacts, and noise. Despite their efforts to learn task-specific features without relying on annotations, these methods often demonstrate a bias towards the training data, which can lead to reduced performance when confronted with new samples, especially in the presence of domain shifts. The absence of labeled data for fine-tuning or retraining the model on the target domain hampers its ability to adapt and generalize effectively. Hence, the performance of these annotation-free methods may suffer when faced with variations in data distribution, making them less robust in scenarios characterized by domain shifts. Therefore, it is crucial to address this limitation and explore techniques that can enhance the adaptability and generalization capabilities of the method.

To address the concern raised by SSL approaches, recent work by Ahn et al. \cite{ahn2021spatial} introduces SGSCN, which specifically designed for segmenting medical images on a single image. This method utilizes different loss functions to facilitate the grouping of spatially connected image pixels with similar feature representations. During the process of iterative learning, the network simultaneously learns feature representations and clustering assignments for each pixel from a single image. Moreover, a contextual aware consistency loss is introduced to enhance the delineation of image regions by enforcing spatial proximity between pixels belonging to the same cluster and its center. More recently, Karimi et al. \cite{karimijafarbigloo2023ms} proposed a novel dual-branch Transformer network. This network aims to capture global contextual dependencies and local information at different scales. Their self-supervised learning approach takes into account the semantic dependency between scales, generating a supervisory signal to ensure inter-scale consistency and enforcing a spatial stability loss within each scale to facilitate content clustering. Additionally, they introduce a cross-entropy loss function applied to the clustering score map, effectively modeling cluster distributions and improving the decision boundary between clusters. Through iterative training, the algorithm progressively learns to assign each pixel to a cluster that is semantically related to it. Building upon this advancement, our method combines a contrastive learning schema with SSL losses to perform accurate and robust semantic segmentation on a single image.

\section{Proposed Method} 
The framework of our proposed method is depicted in \autoref{fig:Model}. The \textbf{S$^3$-Net} integrates local and long-range dependencies to perform the feature embedding process. By incorporating these dependencies, we aim to capture comprehensive contextual information while retaining the local details inherent in the input data.
To achieve effective content clustering without the need for manual annotations, our approach incorporates auxiliary modules and carefully designed data-driven loss functions. These components synergistically facilitate the learning process and promote the formation of meaningful clusters within the embedded feature space. By leveraging the inherent structure and relationships present in the data, our approach empowers the model to drive the segmentation task with robustness and accuracy. In the subsequent subsections, we will provide detailed explanations of each module integrated into our approach. 

\subsection{Encoder Network}
Our encoder architecture comprises three blocks that collectively encode the input image into the latent space. The first block adopts a sequential structure comprising of a $3 \times 3$ convolutional layer, followed by a batch normalization layer. This configuration facilitates the embedding of the input image into a high-dimensional space. In the subsequent stacked $N$ blocks, we employ the I-LKA module to capture both local and global dependencies. To integrate localized descriptions with the I-LKA module's output, a skip connection path is included, followed by another $1 \times 1$ convolutional layer. This setup ensures the preservation of fine-grained spatial information at the pixel level through the skip connection, while simultaneously guiding the network to capture global dependencies using the I-LKA module.
In the last block, we incorporate a deformable convolution layer, specifically designed to effectively model deformations in the lesion boundary, which is a common occurrence in medical images. This additional layer enhances the network's capability to accurately capture and represent deformations, contributing to the overall performance.

\subsubsection{Inception Large Kernel Attention (I-LKA)}
The attention mechanism, the process of identifying the most informative features or important regions, is a critical step toward learning effective feature representation. Two popular attention mechanisms are the self-attention mechanism and the large kernel convolution mechanism, each with its own advantages and drawbacks \cite{azad2023advances}. While the self-attention mechanism excels at capturing long-range dependencies, it lacks adaptability to different channels, exhibits high computational complexity for high-resolution images, and disregards the 2D structure of images. On the other hand, large kernel convolution is effective at establishing relevance and generating attention maps, but it introduces computational overhead and increases parameter count.

To overcome these limitations and leverage the strengths of both self-attention and large kernel convolutions, we propose an enhanced approach called the large kernel attention (LKA) with inception design. In our study, we enhance the LKA module \cite{guo2022visual} by integrating inception strategies. The rationale behind this enhancement is to efficiently capture and integrate information at various spatial resolutions, which is particularly crucial for dense prediction tasks. Unlike the original LKA, which employs fixed-sized filters and thus struggles to fully capture information at different scales within an image, our I-LKA module employs parallel filters of varying sizes to capture both fine-grained details and high-level semantic information concurrently.
The LKA module decomposes a $C \times C$ convolution into three components: a $[\frac{c}{d}]\times [\frac{c}{d}]$ depth-wise dilation convolution ($DW\text{-}D\text{-}Conv$) (representing spatial long-range convolution), a $(2d-1)\times(2d-1)$ depth-wise convolution ($DW\text{-}Conv$) (representing spatial local convolution), and a $1\times1$ convolution (representing channel convolution). This decomposition enables the extraction of long-range relationships within the feature space while maintaining low computational complexity and parameter count when generating the attention map. Our I-LKA define as:
\begin{equation}
    \text{Inc(x)} = \left \{(\mathrm{DW\text{-}Conv}(\mathrm{{F(x)}}))_r | r \in \mathbb{N} \right \}
\end{equation}
\begin{equation}
\text { Attention }=\operatorname{Conv}_{1 \times 1}(\mathrm{DW\text{-}D\text{-}Conv}(Inc(x))
\end{equation}
\begin{equation}
\text { Output }=\text { Attention } \otimes \mathrm{{F(x)}}
\end{equation}
where $Inc(x)$ and $F(x) \in {R}^{C \times H \times W}$ show the inception features and convolutional operation, respectively, while $Attention \in {R}^{C \times H \times W}$ represents the attention map. The symbol $\otimes$ denotes the element-wise product, with the value of the attention map indicating the importance of each feature. Notably, unlike conventional attention methods, the I-LKA approach does not require additional normalization functions such as Sigmoid or SoftMax. 

\subsection{Network Prediction}
Given an input image $X^{H \times W \times C}$, where $H \times W$ represents the spatial dimensions and $C$ denotes the number of channels, our network initiates the segmentation process by utilizing the encoder module to generate a soft prediction map $S^{H \times W \times K}$, where $K$ represents the number of clusters. To obtain the final semantic segmentation map $Y^{H \times W \times K}$, we apply the ArgMax function at each spatial location to activate the corresponding cluster index. During the training phase, we employ an iterative approach to minimize the cross-entropy loss, which measures the discrepancy between the soft prediction map and the segmentation map. By optimizing this loss function, our network learns to produce accurate and meaningful segmentation results:
\begin{equation}
\mathcal{L}_{c e}\left(\mathbf{S}, \mathbf{Y}\right)=-\frac{1}{H \times W} \sum_{i=1}^{H \times W} \sum_{j=1}^K \mathbf{Y}_{i, j} \log \left(\mathbf{S}_{i, j}\right).
\end{equation}
While the cross-entropy loss employed in our approach effectively captures the distribution of clusters by promoting the grouping of similar pixels and increasing the separation between different clusters, it falls short in modeling the spatial relationships within local regions. Consequently, it exhibits limitations in accurately merging neighboring clusters, leading to sub-optimal performance. To address this issue, we propose the integration of a spatial consistency loss, which serves as an additional regularization term. 

\subsection{Spatial Consistency Loss}
\begin{figure}[t]
	\centering
	\includegraphics[width=1\linewidth]{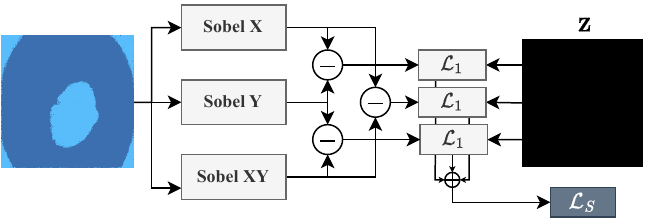}
	\caption{Illustration of the spatial loss calculation process. The spatial loss, denoted by $\mathcal{L}_{\text{S}}$, is computed by subtracting pixel values in $X$, $Y$, and $XY$ directions, and then taking the absolute difference between the resultant image and the zero image $Z$. The summation of these absolute differences is the output of $\mathcal{L}_{\text{S}}$ loss.} 

\label{fig:sptial}
\end{figure}

In addition to the cross-entropy loss, which primarily focuses on capturing the distributional differences among clusters, we introduce the spatial consistency loss to address the limitation of disregarding spatial arrangements. The spatial consistency loss takes into account the local discrepancies in the image by leveraging edge information. By calculating the edges in the $X$, $Y$, and $XY$ directions using the Sobel operator, we can model the spatial relationships and boundaries between regions. By minimizing the pairwise differences based on the edge information, our spatial consistency loss promotes spatial coherence and encourages neighboring pixels with similar visual characteristics to be grouped. This enables our method to not only capture the distributional information but also consider the spatial arrangement of pixels, resulting in more accurate and visually coherent segmentation. The spatial loss function, denoted as $\mathcal{L}_{\text{S}}$, is formulated as follows:
\begin{equation}
\begin{split}
        \mathcal{L}_{\text{S}} = \sum_{i,j} (&\left| (\mathbf{X}_{i,j} - \mathbf{Y}_{i,j}) - \mathbf{Z}_{i,j} \right| + \\ 
        &\left| (\mathbf{X}_{i,j} - \mathbf{XY}_{i,j}) - \mathbf{Z}_{i,j} \right| + \\
        &\left| (\mathbf{Y}_{i,j} - \mathbf{XY}_{i,j}) - \mathbf{Z}_{i,j} \right|),
\end{split}
\end{equation}
where $\mathbf{X}_{i,j}$, $\mathbf{Y}_{i,j}$, and $\mathbf{XY}_{i,j}$ represent the edge information at pixel location $(i,j)$ in the $X$, $Y$, and $XY$ directions, respectively. $\mathbf{Z}_{i,j}$ also represents a zero image with the same dimension as edge images. The $L_{1}$ distance is computed between the pairwise differences of edges in different directions and the zero image, and the absolute differences are summed over all pixels in the image. The goal is to minimize spatial loss, which encourages the alignment of edges and promotes spatial consistency between neighboring pixels. This helps enhance the accuracy and visual coherence of the segmentation results.
The overall process of the spatial consistency loss is depicted in \autoref{fig:sptial}. 
\subsection{Surrogate Task}

To further promote the network's robustness against affine transformations, we introduce an additional segmentation head and leverage the concept of consistency in the feature space. By incorporating affine transformations in both the feature space and the ground truth masks (we use the prediction of the main path as a pseudo label), we aim to establish consistency between the transformed features and the corresponding transformed masks. This approach enables the network to learn positive pairs in the form of contrastive learning, facilitating the generation of a feature space that is resilient to affine transformations.
The motivation behind incorporating consistency over affine transformations lies in the inherent challenges posed by geometric distortions commonly encountered in medical imaging. Affine transformations, such as translation, rotation, scaling, and shearing, can significantly alter the spatial arrangement and appearance of anatomical structures in medical images. Consequently, accurate segmentation in the presence of such transformations becomes a critical requirement for reliable medical image analysis. To this end, we define the affine loss as:
\begin{equation}
\mathcal{L}_{\text{AT}}\left(\mathbf{Y^{'}}, \mathbf{Y^{a}}\right)=-\frac{1}{H \times W} \sum_{i=1}^{H \times W} \sum_{j=1}^K \mathbf{Y^{a}}_{i, j} \log \left({\mathbf{Y^{'}}}_{i, j}\right),
\end{equation}
where $\mathbf{Y^{a}} = \mathbf{A} \cdot \mathbf{Y} + \mathbf{t}$ shows affine transformation (indicated with affinity matrix $\mathbf{A}$ and translation parameter $t$) applied on the network prediction map $\mathbf{Y}$ and $\mathbf{Y}^{'}$ indicate the prediction map of the second path. This consistency over transformation loss encourages the network to minimize the discrepancy between the predicted masks generated from the transformed features and the transformed ground truth masks. This drives the network to become more robust and capable of accurately modeling and segmenting anatomical structures despite variations caused by transformations.

\subsection{Joint Objective}
The final loss function employed in our training process encompasses three distinct loss terms as:
\begin{equation} \label{losses} \mathcal{L}_{\text {joint}}=\lambda_1 \mathcal{L}_{ce}+\lambda_2 \mathcal{L}_{AT}+\lambda_3 \mathcal{L}_{S},
\end{equation}
where, the first term, denoted as $\mathcal{L}_{\text {ce}}$, represents the cross-entropy loss. This term measures the discrepancy between the predicted scores generated by the network and the corresponding maximum index of the ground truth labels. Its purpose is twofold: to ensure prediction confidence by optimizing the agreement between the network's output and the true labels, and to enable the network to learn the distribution characteristics of each cluster.
The second term in our loss function, denoted as $\mathcal{L}_{\text {AT}}$, is designed to enhance the network's invariance against affine transformations. In our self-supervised strategy, we integrate a contrastive learning approach that emphasizes the acquisition of invariance to affine transformations.
The final term, denoted as $\mathcal{L}_{\text {S}}$, is included to promote spatial consistency within each image region. This term aims to reduce local variations and facilitate the smooth merging of neighboring clusters.
To control the relative importance of each loss term, we introduce weighting factors $\lambda_1$, $\lambda_2$, and $\lambda_3$. These factors allow us to balance the influence of each term.

\section{Experiments}
\noindent\textbf{Skin Lesion Segmentation:}
For the first task, we focused on segmenting skin lesion regions in dermoscopic images. To evaluate our method, we utilized the PH$^2$ dataset \cite{mendoncca2013ph}, which consists of 200 RGB images of melanocytic lesions. This dataset offers a diverse range of lesion types, presenting a challenging real-world problem for segmentation. Similar to \cite{ahn2021spatial}, we utilized all 200 samples from the dataset to assess the performance of our method.

\noindent\textbf{Lung Segmentation:}
In the second task, we addressed lung segmentation in CT images. To conduct this evaluation, we employed the publicly available lung analysis dataset provided by Kaggle, as described in \cite{azad2019bi}. This dataset includes both 2D and 3D CT images. Following the approach outlined in \cite{azad2019bi}, we prepared the dataset for evaluation. Specifically, we follow \cite{karimijafarbigloo2023ms} and extract 2D slices from the 3D images, and selected the first 700 samples for our evaluation.
As the organ tissue is usually separable based on the pixel values in this experiment, we have also included the pixel values alongside the score map to predict the lung organ.

\subsection{Experimental Setup}
\noindent\textbf{Training}: To learn the trainable parameters, we employ SGD optimization, minimizing the overall loss function iteratively for a maximum of 50 iterations. The SGD optimization is configured with a learning rate of 0.36 and a momentum of 0.9. The experiments are performed using the PyTorch library on a single RTX 3090 GPU.

\noindent\textbf{Evaluation Protocol}:
We employ the Dice (DSC) score, XOR metric, and Hammoud distance (HM) as evaluation metrics. These metrics allow us to compare our method against both the unsupervised \textit{k}-means clustering method and recent self-supervised approaches, namely DeepCluster \cite{caron2018deep}, IIC \cite{ji2019invariant}, and spatial guided self-supervised strategy (SGSCN) \cite{ahn2021spatial}, and MS-Former \cite{karimijafarbigloo2023ms}. In accordance with the methodology presented in \cite{ahn2021spatial}, we only consider the cluster that exhibits the highest overlap with the ground truth (GT) map as the target class prediction for evaluating our method.
In our evaluation, the DSC score serves as an indicator of the agreement between the predicted target region and the GT map. Higher DSC scores reflect improved performance. Conversely, the HM and XOR metrics measure the discrepancy between the predicted target and the GT map. Therefore, lower HM and XOR values correspond to superior performance.

\begin{figure*}[t]
    \centering
    \resizebox{\textwidth}{!}{
    \begin{tabular}{@{} *{7}c @{}}
    \includegraphics[width=0.20\textwidth]{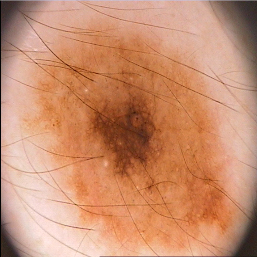} &
    \includegraphics[width=0.20\textwidth]{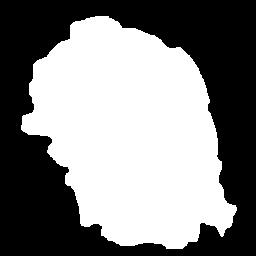} &
    \includegraphics[width=0.20\textwidth]{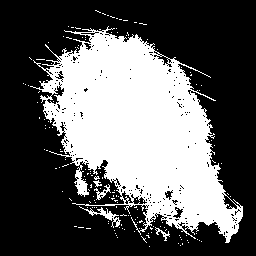} &
    \includegraphics[width=0.20\textwidth]{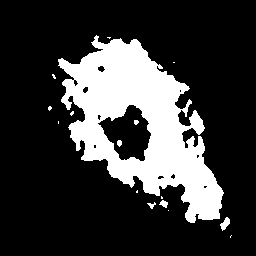} &
    \includegraphics[width=0.20\textwidth]{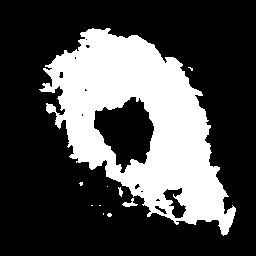} &
    \includegraphics[width=0.20\textwidth]{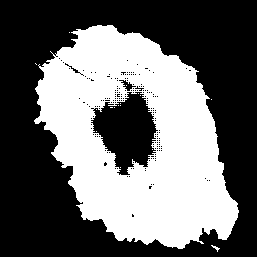} &
    \includegraphics[width=0.20\textwidth]{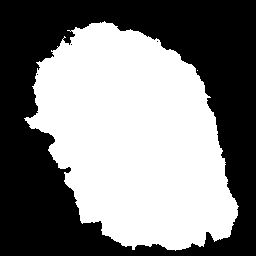} \\
        \includegraphics[width=0.20\textwidth]{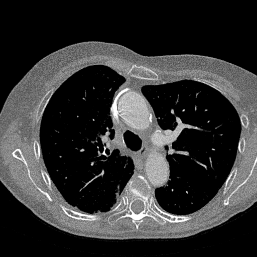} &
    \includegraphics[width=0.20\textwidth]{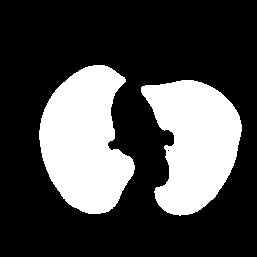} &
    \includegraphics[width=0.20\textwidth]{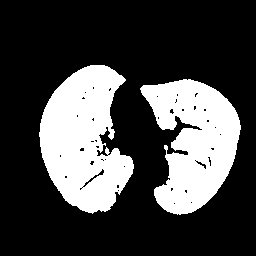} &
    \includegraphics[width=0.20\textwidth]{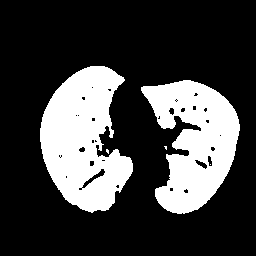} &
    \includegraphics[width=0.20\textwidth]{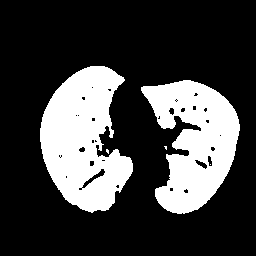} &
    \includegraphics[width=0.20\textwidth]{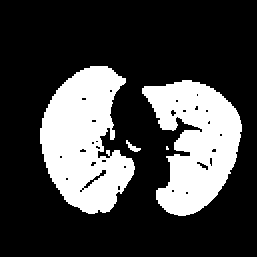} &
    \includegraphics[width=0.20\textwidth]{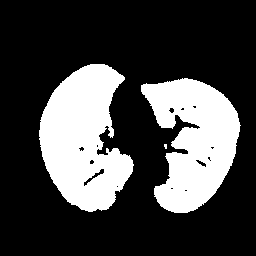} \\
    { \small (a) Input Image} & {\small (b) Ground Truth} & {\small (c) \textit{k}-means} & {\small (d) DeepCluster} & {\small (e) SGSCN} & {\small (f) MS-Former} & {\small (g) Our Method}
    \end{tabular}
} 
    \caption{Visual comparison of different methods on the PH$^2$ skin lesion segmentation and Lung datasets.} 
    \label{fig:skin}

\end{figure*}

\subsection{Evaluation Results}
% In this section, we present the outcome of our suggested method and provide a comprehensive analysis of its performance compared to the SOTA approaches in the tasks of skin lesion segmentation and lung region segmentation. The evaluation includes both quantitative and qualitative assessments, shedding light on the superior capabilities of our strategy. Moreover, we delve into the underlying reasons behind the success of our method, offering valuable insights for further research and development.
\noindent\textbf{Skin Lesion Segmentation}
In the skin lesion segmentation task (refer to \autoref{tab:results}), our method outperforms the SOTA approaches across all evaluation metrics, demonstrating the effectiveness of our self-supervised content clustering strategy. Notably, our method exhibits superior performance compared to SGSCN and MS-Former by modeling spatial consistency at both the pixel and region levels. This modeling of spatial dependency provides a stronger foundation for accurate segmentation. Furthermore, our approach incorporates consistency over transformations, allowing the network to learn transformation-invariant feature representations, leading to smoother clustering space. This recalibration of feature representations contributes to improved segmentation accuracy. The visual comparison in \autoref{fig:skin} confirms the superiority of our strategy, as it generates smoother segmentation maps with better delineation of lesion boundaries compared to DeepCluster and $\textit{K}$-means methods. Additionally, our method successfully avoids under-segmentation issues encountered by MS-Former and SGSCN, where edges around the lesion and inside the lesion class are mistakenly treated as a separate class.

\noindent\textbf{Lung Organ Segmentation}
The quantitative results for lung organ segmentation (refer to \autoref{tab:results}) further highlight the superiority of our self-supervised method over SOTA approaches. 
Our approach demonstrates exceptional performance in addressing the specific challenges encountered when working with CT images, as evidenced by its outstanding results across various evaluation metrics. CT images are known for their inherent noise and the presence of spiky ground truth labels, which can pose significant obstacles to traditional self-supervised methods. However, our self-supervised approach excels in overcoming these challenges and achieves highly accurate segmentation results. Notably, the utilization of the \textit{k}-means algorithm proves particularly effective in lung segmentation tasks, benefiting from the comparatively simpler shapes and lower variations observed in localized areas of the lung dataset. The visual segmentation outputs showcased in \autoref{fig:skin} further validate the efficacy of our approach, as they exhibit noticeably smoother contour lines compared to alternative methods, affirming the superiority of our proposed methodology.
This improvement indicates that the integration of the I-LKA module facilitates the network's ability to accurately perceive the actual boundary of the target.

\begin{table}[h]
    \caption{The performance of the proposed method is compared to the SOTA approaches on the PH$^2$ and Lung datasets.}\label{tab:results}
    
    \centering
    \resizebox{0.49\textwidth}{!}{
    \arrayrulecolor{black}
    \begin{tabular}{c||ccc||ccc} 
    \toprule
    \multirow{2}{*}{\textbf{Methods}} & \multicolumn{3}{c||}{\textbf{PH$^2$}} & \multicolumn{3}{c}{\textbf{Lung Segmentation}} \\ 
    \cline{2-7}
     & \textbf{DSC~$\uparrow$} & \textbf{HM~$\downarrow$} & \textbf{XOR~$\downarrow$} & \textbf{DSC~$\uparrow$} & \textbf{HM~$\downarrow$} & \textbf{XOR~$\downarrow$} \\ 
    \hline
    \textit{k}-means & 71.3 & 130.8 & 41.3 & 92.7 & 10.6 & 12.6 \\
    DeepCluster \cite{caron2018deep} & 79.6 & 35.8 & 31.3 & 87.5 & 16.1 & 18.8 \\
    IIC \cite{ji2019invariant} & 81.2 & 35.3 & 29.8 & - & - & - \\
    SGSCN\cite{ahn2021spatial} & 83.4 & 32.3 & 28.2 & 89.1 & 16.1 & 34.3 \\ 
    MS-Former \cite{karimijafarbigloo2023ms} & 86.0 & 23.1 & 25.9 & 94.6 & \textbf{8.1} & 14.8\\
    \hline
    \rowcolor[rgb]{0.682,0.859,0.855}
    \textbf{Our Method} & \textbf{88.0} & \textbf{20.4} & \textbf{22.0} & \textbf{94.7} & 8.8 & \textbf{13.1}\\
    \bottomrule
    \end{tabular}
    }
    
\end{table}

\begin{figure*}[t]
    \centering
    \resizebox{\textwidth}{!}{
    \begin{tabular}{@{} *{7}c @{}}
    \includegraphics[width=0.20\textwidth]{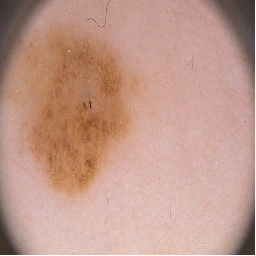} &
    \includegraphics[width=0.20\textwidth]{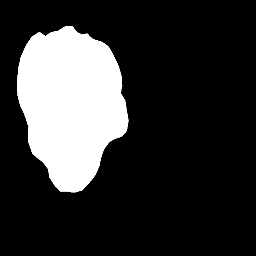} &
    \includegraphics[width=0.20\textwidth]{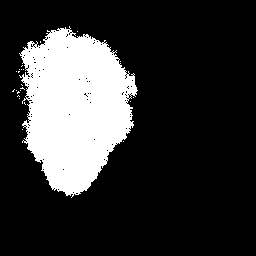} &
    \includegraphics[width=0.20\textwidth]{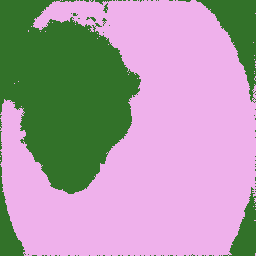} &
    \includegraphics[width=0.20\textwidth]{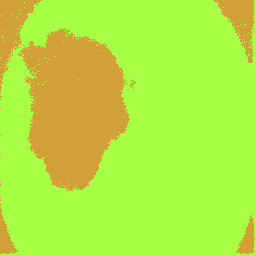} &
    \includegraphics[width=0.20\textwidth]{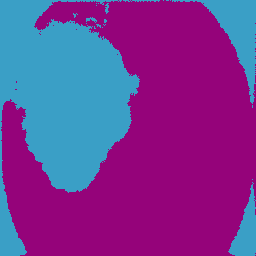} &
    \includegraphics[width=0.20\textwidth]{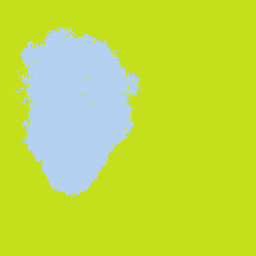} \\
    \includegraphics[width=0.20\textwidth]{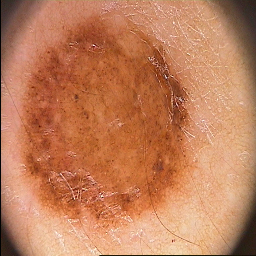} &
    \includegraphics[width=0.20\textwidth]{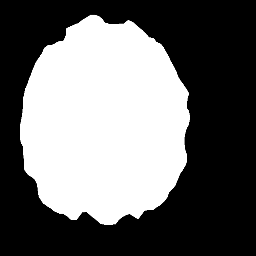} &
    \includegraphics[width=0.20\textwidth]{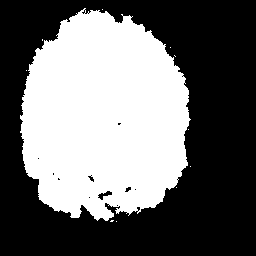} &
    \includegraphics[width=0.20\textwidth]{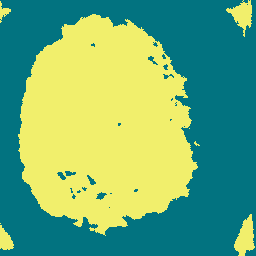} &
    \includegraphics[width=0.20\textwidth]{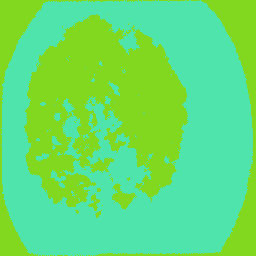} &
    \includegraphics[width=0.20\textwidth]{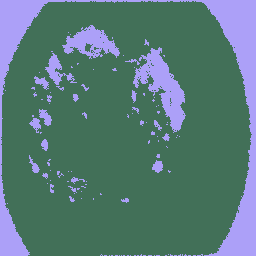} &
    \includegraphics[width=0.20\textwidth]{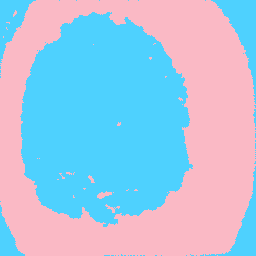} \\
    { \small (a) Input Image} & {\small (b) Ground Truth} & {\small (c) Our Method} & {\small (d) $\mathcal{L}_{ce}$} & {\small (e) $\mathcal{L}_{ce} + \mathcal{L}_{AT}$} & {\small (f) $\mathcal{L}_{ce} + \mathcal{L}_{S}$} & {\small (g) $\mathcal{L}_{joint}$}
    \end{tabular}
} 
    \caption{Segmentation results of the proposed method on the skin lesion segmentation task using the PH$^2$ dataset.} 
    \label{fig:losses}
    
\end{figure*}

\subsection{Ablation Study}
In our proposed method, we integrated two pivotal modules, namely affine transformation invariance and spatial consistency, to improve the feature representation for pixel-level image clustering. This section focuses on conducting an ablation study to explore the process of hyperparameter selection for the loss functions. Furthermore, we examine the influence of these modules on the model's generalization performance.

\noindent\textbf{Hyper-parameter Tuning:}
The hyperparameters for our method were carefully selected and fine-tuned based on empirical evaluations using a small subset of skin lesion segmentation images (10 samples) from the ISIC 2017 dataset \cite{codella2018skin}. We employed a grid search approach within a limited range (0 - 3) to identify the optimal values for $\lambda_1 = 1.2$, $\lambda_2 = 0.3$, and $\lambda_3 = 0.3$. These obtained hyperparameters were used for both datasets.

To comprehensively evaluate the impact of hyperparameters on the new dataset, we conducted a series of additional experiments. The primary objective was to determine the optimal hyperparameters specifically tailored to the lung segmentation dataset, utilizing a subset of ten samples for this purpose. Through an iterative process, we identified the values of $\lambda_1 = 1$, $\lambda_2= 0.5$, and $\lambda_3= 0.6$ as the most effective configuration. Subsequently, we assessed the performance of the model using these updated hyperparameters and observed a slight improvement compared to the original configuration, with a 0.5\% increase in the DSC.

\noindent\textbf{Impact of Affine Transformation:}
Our network architecture incorporates a second branch dedicated to modeling robustness against affine transformations and providing a supervisory signal to learn invariant feature representations. To evaluate the specific contribution of this module, we conducted an experiment excluding the affine consistency loss. The results are summarized in \autoref{tab:loss}. The omission of the affine consistency loss led to a 2.1\% decrease in the DSC score compared to our main strategy. From a qualitative standpoint, as shown in \autoref{fig:losses}, it can be observed that the absence of this loss function resulted in challenges related to accurate boundary separation. Additionally, the absence weakened the condition of multi-scale feature agreement within the network, leading to the incorrect merging of small clusters with neighboring clusters.

\noindent\textbf{Impact of Spatial Consistency:}
Next, we examined the effect of spatial consistency loss on the clustering process. Quantitative results are presented in \autoref{tab:loss}, where it is evident that our model without the spatial consistency loss performed poorly across various metrics. This observation underscores the significance of spatial consistency for segmentation purposes. Notably, our spatial consistency approach incorporates edge information in both vertical and horizontal directions to effectively model local consistency. Moreover, visual evidence presented in \autoref{fig:losses} shows that the absence of spatial consistency resulted in non-consistent cluster predictions and hindered cluster merging. This effect becomes more pronounced when the algorithm deals with complex surfaces.

\begin{table}[h]
    \centering
    \caption{Impact of individual loss functions on model performance. The experiments were conducted using the PH$^2$ dataset.}
    \begin{tabular}{ccc|ccc}
    \textbf{$\mathcal{L}_{\text {ce}}$} & \textbf{$\mathcal{L}_{\text {AT}}$} & \textbf{$\mathcal{L}_{\text {S}}$} & \textbf{DSC~$\uparrow$} & \textbf{HM~$\downarrow$} & \textbf{XOR~$\downarrow$} \\ 
    \hline
     \checkmark & \xmark & \xmark & 86.1 & 22.8 & 25.6 \\
     \checkmark & \checkmark & \xmark & 86.4 & 22.2 & 24.6 \\
     \checkmark & \xmark & \checkmark & 85.9 & 22.7 & 25.2 \\
     \checkmark & \checkmark & \checkmark & \textbf{88.0} & \textbf{20.4} & \textbf{22.0}
     \label{tab:loss}
    \end{tabular}
    
\end{table}

\begin{figure}[h]
    \centering
    \resizebox{\linewidth}{!}{
    \begin{tabular}{@{} *{4}c @{}}
    \includegraphics[width=0.20\textwidth]{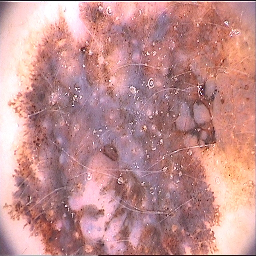} &
    \includegraphics[width=0.20\textwidth]{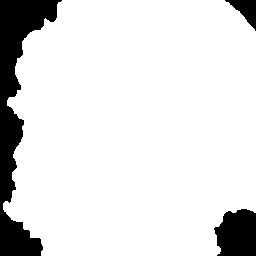} &
    \includegraphics[width=0.20\textwidth]{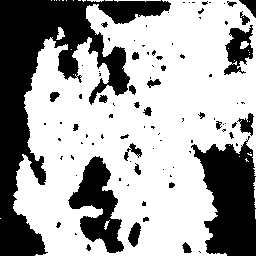} &
    \includegraphics[width=0.20\textwidth]{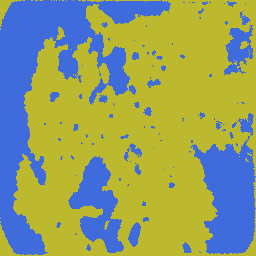} \\
    \includegraphics[width=0.20\textwidth]{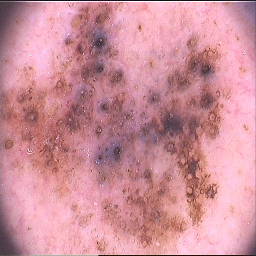} &
    \includegraphics[width=0.20\textwidth]{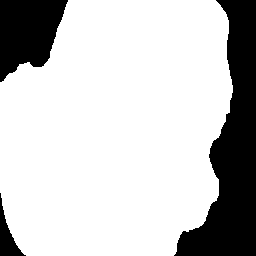} &
    \includegraphics[width=0.20\textwidth]{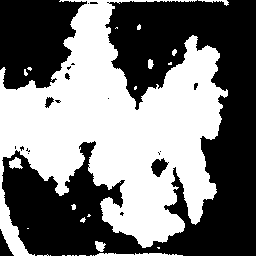} &
    \includegraphics[width=0.20\textwidth]{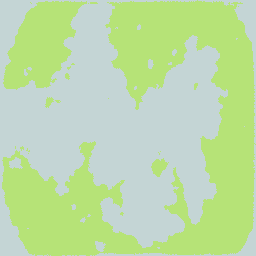} \\
    { \small (a) Input Image} & {\small (b) Ground Truth} & {\small (c) Binary Mask} & {\small (d) Multi-class Mask}
    \end{tabular}
} 
    \caption{Limitations of the proposed method on the PH$^2$ dataset.} 
    \label{fig:limitation}
    
\end{figure}

\noindent\textbf{Impact of Inception and Deformable convolution:}
In order to assess the impact of the inception module in our LKA and the deformable convolution in our architecture, we conducted an experimental analysis by excluding each of these modules individually and replacing them with a simple convolution block. Our results on the PH$^2$ dataset showed a slight decrease of 0.6\% in the DSC score when the inception module was removed. Similarly, the absence of the deformable convolution led to a 0.8\% DSC reduction in performance. These results showcase the importance of both the inception module and the deformable convolution.

\noindent\textbf{Limitations:}
Our proposed method has consistently outperformed SOTA approaches on both datasets, as evidenced by the experimental results. To gain deeper insights into the efficacy of our self-supervised segmentation strategy and to identify potential challenges, we have conducted additional visualizations.
\autoref{fig:limitation} illustrates cases where our proposed method struggles to accurately predict regions of interest, particularly when there is a significant overlap between the object of interest and background regions. 
This limitation becomes more prominent when dealing with complex clustering scenarios, which poses challenges for the model to precisely locate the boundaries of the lesions. Additionally, the presence of noisy annotations in the ground truth masks further impedes the model's ability to generate accurate segmentation maps.

\section{Conclusion}
This paper introduces a novel SSL appraoch that combines the I-LKA module with deformable convolution to enable semantic segmentation directly from the image itself. Additionally, our network incorporates invariance to affine transformations and spatial consistency, providing a promising solution for pixel-wise image content clustering. Experimental results along with the ablation study demonstrate the remarkable performance of our method for skin lesion and organ segmentation tasks.

\noindent\textbf{Acknowledgments}
This work was funded by by the German Research Foundation (Deutsche Forschungsgemeinschaft, DFG) – project number 455548460.

{\small
\bibliographystyle{ieee_fullname}
\bibliography{ref.bib}
}

\end{document}